%% file: main.tex
\documentclass{article}

\usepackage{arxiv}
\usepackage{xcolor}

\usepackage[utf8]{inputenc} 
\usepackage[T1]{fontenc}    
\usepackage{hyperref}       
\usepackage{url}            
\usepackage{booktabs}       
\usepackage{amsfonts}       
\usepackage{nicefrac}       
\usepackage{microtype}      
\usepackage{lipsum}
\usepackage{times}
\usepackage{latexsym}
\usepackage{amssymb}
\usepackage{amsthm}
\usepackage{amsmath}
\usepackage{graphicx}
\usepackage{multirow}
\usepackage{xcolor}
\usepackage{subcaption}
\usepackage[square,sort,comma,numbers]{natbib}
\usepackage[titlenumbered,ruled]{algorithm2e}
\graphicspath{ {figs/} }
\usepackage{url}
\usepackage{color}

\title{FastFusionNet: \\ 
New State-of-the-Art for DAWNBench SQuAD}
\date{}

\author{
    Felix Wu, Boyi Li, Lequn Wang\\
    Cornell University
   \And
   Ni Lao\\
   SayMosaic Inc.
   \And
   John Blitzer\\
   Google Inc.
   \And
   Kilian Q. Weinberger\\
   Cornell University
}

\input{macro.tex}

\begin{document}
\maketitle

\begin{abstract}
In this technical report, we introduce FastFusionNet, an efficient variant of FusionNet~\citep{huang2018fusionnet}. FusionNet is a high performing reading comprehension architecture, which was designed primarily for maximum retrieval accuracy with less regard towards computational requirements. For FastFusionNets we remove the expensive CoVe layers~\citep{McCann2017CoVe} and substitute the BiLSTMs with far more efficient SRU layers~\citep{lei2017sru}. The resulting architecture 
obtains state-of-the-art results on DAWNBench~\citep{coleman2017dawnbench} while achieving the lowest training and inference time on SQuAD~\citep{rajpurkar2016squad} to-date. The code is available at \url{https://github.com/felixgwu/FastFusionNet}.
\end{abstract}

\section{Introduction}
\input{sections/intro.tex}

\section{Background}
\input{sections/background.tex}

\section{FastFusionNet}
\input{sections/method.tex}

\section{Experiments}

\input{sections/result.tex}

\section*{Acknowledgments}
This research is supported in part by grants from the National
Science Foundation (III-1618134, III-1526012, IIS1149882,
IIS-1724282, and TRIPODS-1740822), the Office
of Naval Research DOD (N00014-17-1-2175), and the
Bill and Melinda Gates Foundation. We are thankful for
generous support by SAP America Inc.

\bibliographystyle{plainnat}  
\bibliography{references}  

\end{document}

%% file: macro.tex
\renewcommand{\vec}[1]{\boldsymbol{\mathbf{#1}}}

\def\bb{\vec{b}}
\def\bc{\vec{c}}

\def\bh{\vec{h}}

\def\bq{\vec{q}}
\def\br{\vec{r}}

\def\bx{\vec{x}}

\def\R{\mathbb{R}}
\def\bisru{\mathrm{BiSRU}}
\def\attn{\mathrm{Attn}}

\def\mq{\vec{Q}}
\def\mc{\vec{C}}

\newcommand{\nil}[1]{}
\newcommand{\fw}[1]{}
\newcommand{\jb}[1]{}
\newcommand{\kqw}[1]{}
\newcommand{\lw}[1]{}
\newcommand{\boyi}[1]{}

%% file: sections/intro.tex
Recently, machine reading comprehension, or question answering, has received a significant amount of attention in the field of natural language processing.
Reading comprehension tasks focus on an agent's ability to read a piece of text and subsequently answer questions about it. An array of reading comprehension datasets have been released in the past years including WikiQA~\cite{yang2015wikiqa}, SQuAD~\citep{rajpurkar2016squad}, SQuAD v2~\citep{rajpurkar2018know}, TriviaQA~\citep{joshi2017trivia}, SearchQA~\citep{dunn2017searchqa}, NarrativeQA~\citep{kovcisky2018narrativeqa}, CoQA~\citep{reddy2018coqa}, QuAC~\citep{choi2018quac}, and Natural Questions~\citep{kwiatkowski2019natural}.
SQuAD is one of the most popular datasets, where a model is presented with a question-context pair and asked to highlight a span in the context as the answer to the question.
\autoref{txt:qa_samples} shows some examples from SQuAD dataset.

\begin{figure}[h]
  \centering
  \fbox{
  \begin{minipage}{0.8\textwidth}
	\textit{Q: How many Americans are richer than more than half of all citizens?}\\
	According to PolitiFact the top \textbf{400} richest Americans "have more wealth than half of all Americans combined." According to ...
	\\\\
    \textit{Q: What philosophy of thought addresses wealth inequality?}\\
    \textbf{Neoclassical economics} views inequalities in the distribution of income as arising from differences in value added by labor, capital and land. Within labor ...
	\\\\
	\textit{Q: What is the term that describes the difference between what higher paid and lower paid professionals earn?} \\
    ... Thus, in a market economy, inequality is a reflection of the \textbf{productivity gap} between highly-paid professions and lower-paid professions.
  \end{minipage}}
  \caption{Question/Answer samples from SQuAD~\citep{rajpurkar2016squad}}
  \label{txt:qa_samples}
\end{figure}

Although automatic reading comprehension systems have recently reached super-human performance by some benchmarks~\citep{devlin2018bert} (with the help of unsupervised pre-training on large-scale datasets), 
less attention has been paid to their  computational efficiency, which is a crucial aspect in the context of training and deploying such models in real world applications.
\citet{coleman2017dawnbench} introduce DAWNBench, a benchmark suite for end-to-end deep learning training and inference. While many teams have shown interest in the image classification tracks, only few (ParlAI and Google) have participated in the question answering tracks.
The ParlAI~\citep{miller2017parlai} team has successfully reduced the training time of the model to $27$ minutes, while maintaining a decent retrieval accuracy (F1 score greater than or equal to $75 \%$).

In this technical report, we analyze the inference bottlenecks of FusionNet~\citep{huang2018fusionnet} and introduce FastFusionNet that tackles them. In our experiments, we show that FastFusionNet achieves new state-of-the-art training and inference time on SQuAD based on the metrics of DAWNBench.

%% file: sections/background.tex
\subsection{Efficient Sequence Encoding}
The sequential nature of Recurrent Neural Networks (RNNs) makes them inherently slow, even on parallel computing devices. Consequently, a series of methods have been proposed to either reduce the sequential computation within RNNs or substitute them with alternative building blocks.
\citet{yu2017learning} propose LSTM-Jump, in which LSTMs~\cite{Hochreiter1997LongSM} are trained to predict the number of tokens to skip.
\citet{seo2017neural} propose skim-RNN for sentiment analysis and question answering, which has a special RNN unit combining big and small RNN cells.
\citet{bradbury2016quasi} introduce Quesi-RNNs that combines convolution with sequential pooling to reduce the sequential.
\citet{lei2017sru} invent Simple Recurrent Unit (SRU) a fast RNN variant, which will be explained further in \autoref{subsec:sru}. Their second version~\citep{lei2018sru} is more accurate but a little less efficient.
To the best of our knowledge, SRU is the most efficient RNN variant, so we choose it as our preferred building block throughout this manuscript.

Other lines of work replace RNNs with convolution layers~\citep{zhang2015character,dauphin2016glu,gehring2017convseq,kaiser2017depthwise,wu2017fast,wei2018qanet,wu2018pay} or self-attention~\citep{vaswani2017attention,shen2018fast,wei2018qanet,shen2018bi}.
\citet{shen2018bi} introduce bidirectional block self-attentions (Bi-BloSA) that split a sequence into blocks and compute intra-block and inter-block self-attention that significantly reduces computation and memory footprint compared to the popular multi-head self-atttention~\citep{vaswani2017attention}.
\citet{wu2018pay} propose lightweight and dynamic convolutions as efficient alternatives to self-attentions with comparable performance.

\subsection{Simple Recurrent Unit}
\label{subsec:sru}
The key idea behind Simple Recurrent Unit (SRU) \citep{lei2017sru,lei2018sru} is to separate the matrix multiplications (the bottleneck) from the recurrence. To be specific, SRU replaces the matrix multiplication style recurrence to a vector summation style recurrence. As a consequence, the matrix multiplication can be done in parallel at once. The complete architecture is 
\begin{align*}
    &\color{blue} \tilde{\bx}_t = \vec{W}\bx_t\\
    &\color{blue} \vec{f}_t = \sigma(\vec{W}_f \bx_t + \bb_f)\\
    &\color{blue} \vec{r}_t = \sigma(\vec{W}_r \bx_t + \bb_r)\\
    & \bc_t = \vec{f}_t \odot \bc_{t-1} + (1 - \vec{f}_t) \odot \tilde{\bx}_t\\
    & \color{purple} \bh_t = \br_t \odot \mathrm{tanh}(\bc_t) + (1 - \vec{r}_t) \odot \bx_t
\end{align*}
where $\bx_t$ is the input at the time step $t$, $\bc_t$ is the hidden state and $\bh_t$ is the output. All {\color{blue} blue} computations can be performed through simple parallel matrix$\times$matrix multiplies followed by parallel element-wise function operators, and are therefore maximally efficient on modern CUDA hardware. The only sequential operation is the update to $\bc_t$, which is a highly efficient vector operation. The final update to $\color{purple}\bh_t$ is again fully parallel.

\subsection{DrQA}
DrQA~\citep{chen2017reading} is one of the simplest reading comprehension model, which employs a variety of features including pre-trained word vectors, term frequencies, part-of-speech tags, name entity relations, and the fact that whether a context word is in the question or not, encodes the features with RNNs, and predicts the start and end of an answer with a PointerNet-like module~\citep{vinyals2015pointer}.

\section{Analysis of FusionNet}
FusionNet~\citep{huang2018fusionnet} is reading comprehension model built on top of DrQA by introducing Fully-aware attention layers (context-question attention and context self-attention), contextual embeddings~\citep{McCann2017CoVe}, and more RNN layers.
Their proposed fully-aware attention mechanism uses the concatenation of layers of hidden representations as the query and the key to compute attention weights, which shares a similar intuition as DenseNet ~\citep{huang2017densely}. FusionNet was the state-of-the-art reading comprehension model at the time of writing (Oct. 4th 2017).

\begin{figure}[]
    \centering
    \begin{minipage}{.45\textwidth}
        \includegraphics[width=\textwidth]{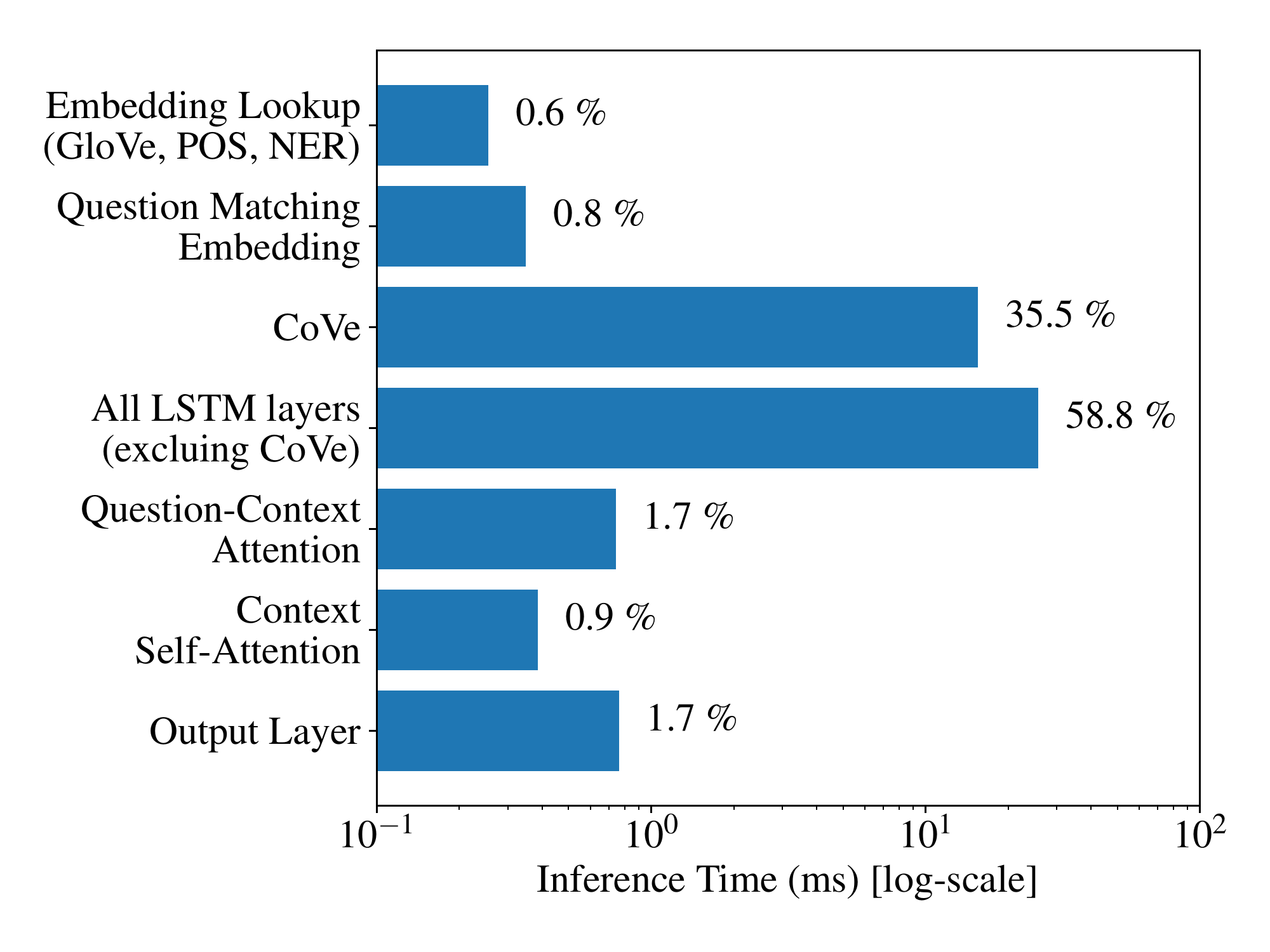}

        \caption{The time spent on different components of FusionNet during inference. Note that we have to block CUDA threads to time each component, which may not reflect the real inference time when we remove the components and forward the whole model without blocking.}
        \label{fig:component}
    \end{minipage}
    \quad
    \begin{minipage}{.45\textwidth}
        \includegraphics[width=\textwidth]{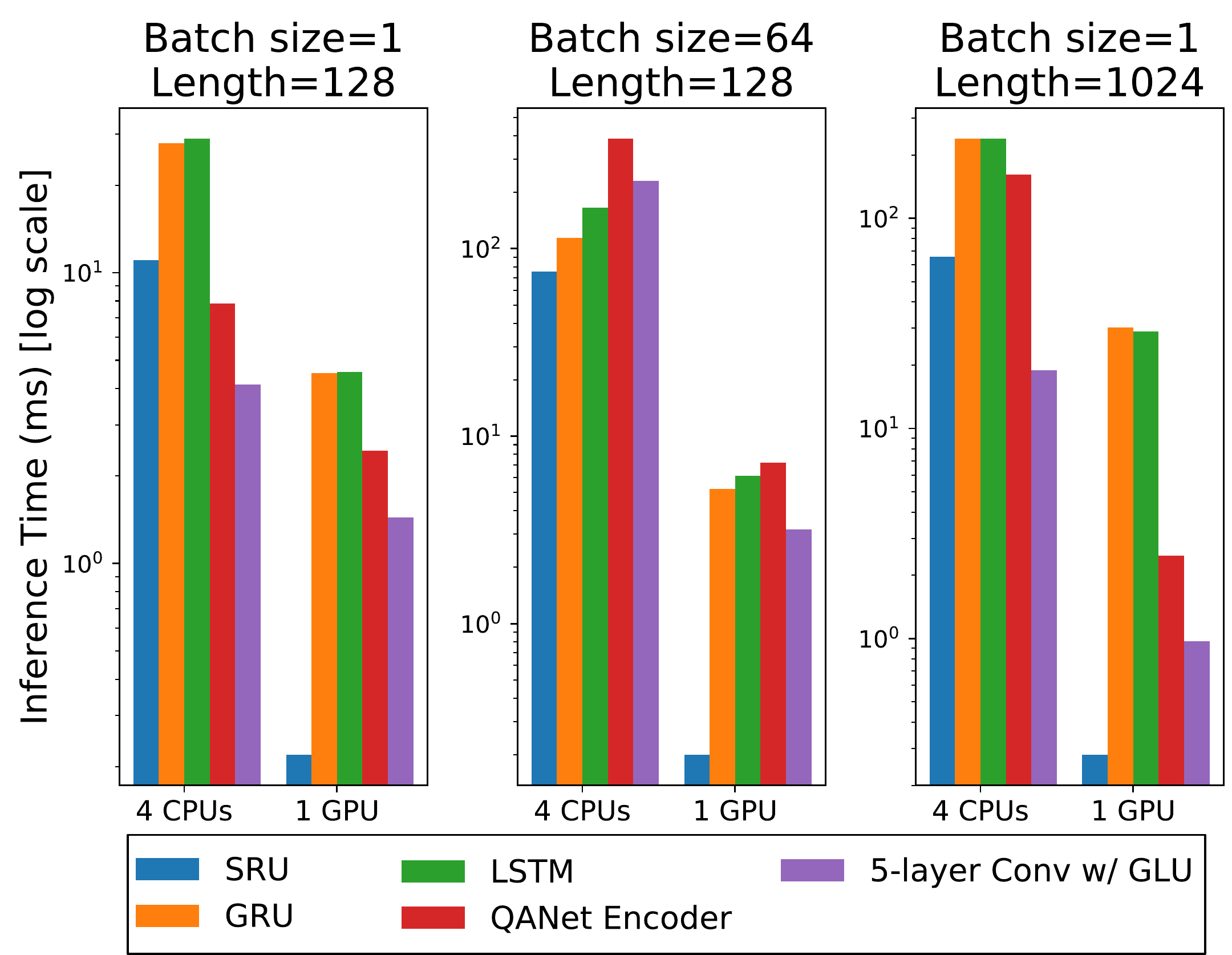}

        \caption{Inference time in log-scale of SRU~\citep{lei2017sru}, GRU~\citep{cho2014gru}, LSTM~\citep{Hochreiter1997LongSM}, QANet Encoding block (with 2 conv layers and a 8-head attention)~\citep{wei2018qanet}, 5 Convolution layers with gated linear unit (GLU)\citep{dauphin2016glu,wu2017fast}. 
        All input and hidden sizes are 128.}
        \label{fig:block_benchmark}
    \end{minipage}
\end{figure}

\autoref{fig:component} provides an analysis of the individual components of FusionNet 
that the contextual embedding layer, i.e. CoVe~\citep{McCann2017CoVe}, with several layers of wide LSTMs, takes up to $35.5\%$ of the inference time while only  contributing a $1.1 \%$ improvement of F1 Score (from $82.5 \%$ to $83.6 \%$)~\citet{huang2018fusionnet}. Additionally, the LSTM layers contribute to $58.8\%$ of the inference time. Therefore, we propose to remove the contextual embedding layer and replace each bidirectional LSTM layer with two layers of bidirectional SRU~\cite{lei2017sru}.
\autoref{fig:block_benchmark} shows that SRU is faster than LSTM~\citep{Hochreiter1997LongSM}, GRU~\citep{cho2014gru}, QANet Encoder~\citep{wei2018qanet}, and 5-layer CNN w/ GLU~\citep{dauphin2016glu,wu2017fast}. We time a 5-layer CNN since it matches the performance of one layer SRU.

%% file: sections/method.tex
Here we introduce FastFusionNet which addresses the inference bottlenecks of FusionNet~\citep{huang2018fusionnet}.
There are two differences compared to FusionNet: i) the CoVe~\citep{McCann2017CoVe} layers are removed and ii) each BiLSTM layer is replaced with two BiSRU layers.

We closely follow the implementation of \citet{huang2018fusionnet} described in their paper except for the changes above. Following \citet{huang2018fusionnet}, the hidden size of each SRU is set to 125, resulting in a 250-d output feature of each BiSRU regardless of the input size.
In the following explanation, we use $[\vec{A}; \vec{B}]$ to represent concatenation in the feature dimension. $\attn(\vec{Q}, \vec{K}, \vec{V})$ represents the attention mechanism taking the query $\vec{Q}$, the key $\vec{K}$, and the value $\vec{V}$ as inputs. Assuming $\vec{O}$ being the output, we have $\vec{O}_i = \sum_{j} \bar{\alpha}_{ij} \vec{V}_j, \bar{\alpha}_{ij} = \frac{\exp(\alpha_{ij})}{\sum_k \exp(\alpha_{ik})}, \alpha_{ij} = \mathrm{ReLU}(\vec{W}\vec{Q})^{\top}\mathrm{ReLU}(\vec{W}\vec{K})$.

\paragraph{Input Features.} Following \citet{chen2017reading}, we use $300$-dim GloVe \citep{pennington2014glove} vectors, term-frequency, part-of-speech (POS) tags, and named entity recognition (NER) tags as features for each word in the context or the question. We fine-tune the embedding vector of the padding token, the unknown word token, and the top $1000$ most frequent words in the training set.  Like others \cite{huang2018fusionnet} we use a randomly initialized the trainable embedding layer with 12 dimensions for POS tags and 8 dimensions for NER. We use question matching features proposed by \citet{chen2017reading} as well, which contains a hard version and a soft version. The hard version contains $3$ binary features indicating where a context word's original form, lower case form, or lemmatized form appears in the question, respectively. The soft version uses a trainable attention module that learns to represent each context word as a mixture of question words. Overall, the $i$-th context token is represented as $\mc_i$ which has $624$ dimensions, and the $j$-th question token is represented as a $300$-d $\mq_j$ glove vector. 
We have the context features $\mc \in \R^{n \times 624}$ and question features $\mq \in \R^{m \times 300}$ where $m$ and $n$ are the length of the question and context, respectively. Specifically,
\begin{align*}
    & \mc^\mathrm{soft\_match} \leftarrow \attn(\mc^\mathrm{GloVe}, \mq^\mathrm{GloVe}, \mq^\mathrm{GloVe})\\
    & \mc^\mathrm{In} \leftarrow [\mc^\mathrm{GloVe}; \mc^\mathrm{TF}; \mc^\mathrm{POS}; \mc^\mathrm{NER}; \mc^\mathrm{soft\_match}; \mc^\mathrm{hard\_match}],\\
    & \mq^\mathrm{In} \leftarrow \mq^\mathrm{GloVe},
\end{align*}
where  $\mc \in \R^{n \times 624}, 
\mc^\mathrm{GloVe} \in \R^{n \times 300},
\mc^\mathrm{TF} \in \R^{n \times 1},
\mc^\mathrm{POS} \in \R^{n \times 12},
\mc^\mathrm{NER} \in \R^{n \times 8},
\mc^\mathrm{hard\_match} \in \R^{n \times 3},$ and 
$\mq = \mq^\mathrm{GloVe} \in \R^{m \times 300}$.

\subparagraph{Low-level encoding Layer.}
We apply 2-layer BiSRU on $\mc^\mathrm{In}$ and $\mq^\mathrm{In}$ to obtain lower-level representations $\mc^\ell $ and $\mq^\ell$ respectively. That is,
\begin{align*}
    & \mc^\ell \leftarrow \bisru (\bisru (\mc^\mathrm{In})),\\
    & \mq^\ell \leftarrow \bisru (\bisru (\mq^\mathrm{In})),
\end{align*}
where  $\mc^\ell \in \R^{n \times 250}, \mq^\ell \in \R^{m \times 250}$.

\subparagraph{High-level Encoding Layer} consists of another  2-layer BiSRU to obtain high-level representations $\mc^h$ and $\mq^h$. In other words,
\begin{align*}
    & \mc^h \leftarrow \bisru (\bisru (\mc^\ell)),\\
    & \mq^h \leftarrow \bisru (\bisru (\mq^\ell)),
\end{align*}
where  $\mc^h \in \R^{n \times 250}, \mq^h \in \R^{m \times 250}$.

\subparagraph{The Question Understanding Layer} is another 2-layer BiSRU combining $\mq^\ell$ and $\mq^h$ into $\mq^u$, i.e.
\[
    \mq^u \leftarrow \bisru (\bisru ([\mq^\ell; \mq^h ])),
\]
where  $\mq^u \in \R^{m \times 250}$.

\subparagraph{The Question-Context Attention Layer} is a fully-aware attention module~\citep{huang2018fusionnet} which takes the history  (concatenation of GloVe, low-level, and high-level features) of each context word and question words as query and key for three attention modules, and represents each context word as three different vectors: $\hat{\mc}^\ell$ (weighted sum of $\mq^\ell$'s), $\hat{\mc}^h$ (weighted sum of $\mq^h$'s), and $\hat{\mc}^u$ (weighted sum of $\mq^u$'s). Another 2-layer SRU processes the concatenation of all previous context word vectors $\mc^\ell, \mc^h, \hat{\mc}^\ell, \hat{\mc}^h$, and $\hat{\mc}^u$ into $\mc^v$. To be specific,
\begin{align*}
    & \mc^\mathrm{His} \leftarrow [\mc^\mathrm{GloVe}; \mc^\mathrm{CoVe}; \mc^\ell; \mc^h],\\
    & \mq^\mathrm{His} \leftarrow [\mq^\mathrm{GloVe}; \mq^\mathrm{CoVe}; \mq^\ell; \mq^h],\\
    & \hat{\mc}^\ell \leftarrow \attn(\mc^\mathrm{His}, \mq^\mathrm{His}, \mq^\ell)\\
    & \hat{\mc}^h \leftarrow \attn(\mc^\mathrm{His}, \mq^\mathrm{His}, \mq^h)\\
    & \hat{\mc}^u \leftarrow \attn(\mc^\mathrm{His}, \mq^\mathrm{His}, \mq^u)\\
    & \mc^v \leftarrow \bisru (\bisru ([\mc^\ell; \mc^h; \hat{\mc}^\ell; \hat{\mc}^h; \hat{\mc}^u])),
\end{align*}
where $\hat{\mc}^\ell, \hat{\mc}^h, \hat{\mc}^u, \mc^v \in \R^{n \times 250}$. 

\subparagraph{The Context Self-Attention Layer} is another fully-aware attention module that treats the history of words (GloVe vectors,  $\mc_i^\ell, \mc_i^h, \hat{\mc}_i, \hat{\mc}_i^h, \hat{\mc}_i^u,$ and $\mc_i^v$) as the key and also as query to produce a new vector of each context word $\hat{\mc}_i^v$ from the input $\mc_i^v$. The last 2-layer SRU processes the concatenation of $\mc_i^v$ and $ \hat{\mc}_i^v $ into $\mc_i^u$, i.e.
\begin{align*}
    & \mc^\mathrm{His2} \leftarrow [\mc^\mathrm{GloVe}; \mc^\mathrm{CoVe}; \mc^\ell; \mc^h; \hat{\mc}^\ell; \hat{\mc}^h; \hat{\mc}^u; \mc^v],\\
    & \hat{\mc}^v \leftarrow \attn(\mc^\mathrm{His2}, \mc^\mathrm{His2}, \mc^v)\\
    & \mc^u \leftarrow \bisru (\bisru ([\mc^v; \hat{\mc}^v])),
\end{align*}
where $\hat{\mc}^v, \mc^u \in \R^{n \times 250}$.

\subparagraph{Answer Prediction Layer.}
This layer predicts the positions of the start and end of the answer span using the final representations of the context $\mc^u$ and the question $\mq^u$.
This layer first combines all question vectors into a weighted sum $\bq = \sum_{j=1}^m \alpha_j \mq_j^u$ using a single trainable parameter $\mathbf{v} \in \R^{250}$, where $\alpha_j = \frac{\exp{(\mathbf{v}^\intercal \mq_j^u})}{\sum_{j=1}^m\exp{(\mathbf{v}^\intercal \mq_j^u)}}$.
As a next step it predicts the probability that the $i^{th}$ word denotes the start of the answer span as
 $s_i = \frac{\exp{(\bq^{\top} W_1 \mc_i^u)}}{\sum_{i=k}^n \exp{(\bq^{\top} W_1 \mc_k^u})}$, 
using a bi-linear soft-max model.   Subsequently, it summarizes the context with the start prediction and produces $\mathbf{z} = \sum_{i=1}^n s_i \mc_i^u$. It then produces a refined  question vector $\hat{\bq}$ with one step of GRU \citep{cho2014gru}, using the original question vector $\bq$ as the hidden memory and $\mathbf{z}$ as the input, i.e. $\hat{\bq} = \mathrm{GRUCell}(\mathbf{z}, \bq)$.
Similarly, a bi-linear module is applied to get the end predicted probability $e_i = \frac{\exp{(\hat{\bq}^{\top} W \mc_i^u)}}{\sum_{k=1}^n \exp{(\hat{\bq}^{\top} W \mc_k^u)}}$.
The product of the respective start and end probabilities becomes the score of an answer span. However, we only consider answers with no more than $15$ words and do a exhaustive search to find the best span.

%% file: sections/result.tex
\subsection{Experimental Setup}
We conduct our experiments on the SQuAD dataset, which contains 87K, 10K, and 10K context-question pairs for training, development, and test. Like the other models submitted to the DAWNBench, we use the publicly available development set to evaluate the performance and the efficiency of our model. 
All of the experiments are conducted on a single Nvidia GTX-1080 Ti GPU. 
We use PyTorch~\citep{paszke2017automatic} 0.3.1 to implement our model.
We use single precision floating-point in our implementation. Arguably, using half-precision floating-point may further improve our results.
Our implementation is based on two open source code base\footnote{\url{https://github.com/hitvoice/DrQA} and \url{https://github.com/momohuang/FusionNet-NLI}}. We follow their data pre-processing procedure.

\paragraph{Training procedure.}
We train the model for 100 epochs to ensure convergence; however, the model stops improving after 60 epochs.
The other hyper-parameters are borrowed from \citet{lei2017sru}.  We do not tune the hyper-parameters. We use batch size $32$ for training. 
We use Adam optimizer \cite{kingma2014adam} with $\alpha = 0.001$ and clip the $\ell_2$-norm of the gradients to $20$ before each update. The SQuAD dataset is tokenized and tagged by the SpaCy package \footnote{\url{https://spacy.io/} We use version 1.9.0}. We apply variational dropout \cite{kingma2015variational} to sequential features and normal dropout \cite{Srivastava2014dropout} to others. Following \cite{chen2017reading}, dropout rate for input embeddings is set to $0.4$. We also dropout all inputs of LSTMs and attentions with probability $0.4$. For SRUs, we follow \cite{lei2017sru} using dropout rate 0.2. We do not use learning rate decay or weight decay for simplicity.

\subsection{DAWNBench Results}
We report the performance of our FastFusionNet on DAWNBench~\citep{coleman2017dawnbench}. We consider three baselines: i) FusionNet ii) FusionNet without CoVE, and iii) BERT-base. For BERT-base, we use the open source code\footnote{\url{https://github.com/google-research/bert}}. Our FastFusionNet reaches F1 $75 \%$ in 4 epochs and achieves at F1 $82.5 \%$ at the end which matches the reported F1 $82.5 \%$ of FusionNet without CoVe on SQuAD development set\citep{huang2018fusionnet}.

\begin{table}[t]
    \centering
    \begin{tabular}{r|l|l|l}
    \toprule
    Time to F1 $\ge$ 75.0\%  & Model  & Framework & Hardware            \\
    \midrule
    \color{blue}\textbf{0:18:46}   & \color{blue}FastFusionNet (4 epochs) & \color{blue} PyTorch v0.3.1  & 1 \color{blue}GTX-1080 Ti \\
    \color{blue}0:23:06   & \color{blue}FusionNet~\citep{huang2018fusionnet} (2 epochs) & \color{blue} PyTorch v0.3.1  & 1 \color{blue}GTX-1080 Ti \\
    0:27:07 & DrQA (ParlAI) \citep{miller2017parlai} & PyTorch v1.0.0 & 1 RTX-2080  \\
    \color{blue}0:29:24   & \color{blue}FusionNet without CoVe (3 epochs)~\citep{huang2018fusionnet} & \color{blue} PyTorch v0.3.1  & 1 \color{blue}GTX-1080 Ti \\
    0:45:56  & QANet \citep{wei2018qanet} & TensorFlow v1.8 & 1 TPUv2  \\  
    0:50:21 & DrQA (ParlAI) \citep{miller2017parlai} & PyTorch v1.0.0 & 1 T4 / GCP \\
    0:56:43 & DrQA (ParlAI) \citep{miller2017parlai} & PyTorch v1.0.0 & 1 P4 / GCP \\
    1:00:35 & DrQA (ParlAI) \citep{miller2017parlai} & PyTorch v0.4.1 & 1 V100 \\
    \color{blue}1:22:33 & \color{blue}BERT-base \citep{devlin2018bert} (1 epoch fine-tuning) & \color{blue} TensorFlow v1.11.0 & \color{blue} 1 GTX-1080 Ti\\
    7:38:10   & BiDAF\cite{seo2016bidirectional, coleman2017dawnbench} & TensorFlow v1.2 & 1 K80  \\
    \bottomrule
    \end{tabular}
    \caption{DAWNBench Training Track}
    \label{tab:dawnbench_train}
\end{table}

\begin{table}[t]
    \centering
    \begin{tabular}{r|l|l|l}
    \toprule
    1-example Latency & Model (F1 $\ge 75\%)$  & Framework   & Hardware\\
    \midrule
    \color{blue}\textbf{7.9 ms}     & \color{blue}FastFusionNet (F1 $82.5\%$) & \color{blue}PyTorch v0.3.1  & \color{blue}1 GTX-1080 Ti \\
    \color{blue}22.3 ms     & \color{blue}BERT-base (F1 $88.5\%$) & \color{blue}TensorFlow v1.11.0 & \color{blue}1 GTX-1080 Ti \\
    \color{blue}{32.6 ms}     & \color{blue}FusionNet without CoVe~\citep{huang2018fusionnet} (F1 $82.5\%^*$) & \color{blue}PyTorch v0.3.1  & \color{blue}1 GTX-1080 Ti \\
    \color{blue}{45.5 ms}     & \color{blue}FusionNet~\citep{huang2018fusionnet} (F1 $83.6\%^*$) & \color{blue}PyTorch v0.3.1  & \color{blue}1 GTX-1080 Ti \\
    100.0 ms            & BiDAF (F1 $77.3\%$)  & TensorFlow v1.2 & 16 CPU    \\
    590.0 ms            & BiDAF (F1 $77.3\%$)  & TensorFlow v1.2 & 1 K80\\
    638.1 ms            & BiDAF (F1 $77.3\%$)  & TensorFlow v1.2 & 1 P100\\
    \bottomrule
    \end{tabular}
    \caption{DAWNBench Inference Track. $^*$: We use the F1 score reported by \citet{huang2018fusionnet} here since our re-implementation is about $0.5\%$ F1 score worse.}
    \label{tab:dawnbench_infer}
\end{table}

\paragraph{The training time track}
aims to minimize the time to train a model up to at least $75 \%$ F1 score on SQuAD development set.
\autoref{tab:dawnbench_train} shows that our FastFusionNet reaches F1 $75.0 \%$ within 20 minutes (after 4 epochs), which gives a $45\%$ speedup compared to the winner DrQA(ParlAI) on the leaderboard. Notably, we use an Nvidia GTX-1080 GPU which is about $22\%$ slower than their Nvidia RTX-2080 GPU. When controlling the generation of GPUs and comparing our model with a DrQA (ParlAI) trained on an Nvidia V100, our model achieves a $3.1\times$ speedup. Compared to FusionNet, FastFusionNet is $23\%$ faster to reach $75\%$ F1 score; however, in terms of the training time per epoch, it is in fact $2.6\times$ as fast as FusionNet.

\paragraph{The inference time track} evaluates the average 1-example inference latency of a model with an F1 score at least $75\%$. Our FastFusionNet reduces the 1-example latency down to 7.9 ms, which is $2.8\times$ as fast as a BERT-base and $12.7\times$ over BiDAF.
FastFusionNet achieves a $5.8\times$ speedup over the original FusionNet.

%% file: main.bbl
\begin{thebibliography}{40}
\providecommand{\natexlab}[1]{#1}
\providecommand{\url}[1]{\texttt{#1}}
\expandafter\ifx\csname urlstyle\endcsname\relax
  \providecommand{\doi}[1]{doi: #1}\else
  \providecommand{\doi}{doi: \begingroup \urlstyle{rm}\Url}\fi

\bibitem[Bradbury et~al.(2017)Bradbury, Merity, Xiong, and
  Socher]{bradbury2016quasi}
James Bradbury, Stephen Merity, Caiming Xiong, and Richard Socher.
\newblock {Quasi-Recurrent Neural Networks}.
\newblock In \emph{International Conference on Learning Representations}, 2017.

\bibitem[Chen et~al.(2017)Chen, Fisch, Weston, and Bordes]{chen2017reading}
Danqi Chen, Adam Fisch, Jason Weston, and Antoine Bordes.
\newblock Reading {Wikipedia} to answer open-domain questions.
\newblock In \emph{Association for Computational Linguistics}, 2017.

\bibitem[Cho et~al.(2014)Cho, van Merrienboer, Gulcehre, Bahdanau, Bougares,
  Schwenk, and Bengio]{cho2014gru}
Kyunghyun Cho, Bart van Merrienboer, Caglar Gulcehre, Dzmitry Bahdanau, Fethi
  Bougares, Holger Schwenk, and Yoshua Bengio.
\newblock Learning phrase representations using rnn encoder--decoder for
  statistical machine translation.
\newblock In \emph{Empirical Methods in Natural Language Processing}, 2014.

\bibitem[Choi et~al.(2018)Choi, He, Iyyer, Yatskar, Yih, Choi, Liang, and
  Zettlemoyer]{choi2018quac}
Eunsol Choi, He~He, Mohit Iyyer, Mark Yatskar, Wen-tau Yih, Yejin Choi, Percy
  Liang, and Luke Zettlemoyer.
\newblock Quac: Question answering in context.
\newblock In \emph{Proceedings of the 2018 Conference on Empirical Methods in
  Natural Language Processing}, pages 2174--2184, 2018.

\bibitem[Coleman et~al.(2017)Coleman, Narayanan, Kang, Zhao, Zhang, Nardi,
  Bailis, Olukotun, R{\'e}, and Zaharia]{coleman2017dawnbench}
Cody Coleman, Deepak Narayanan, Daniel Kang, Tian Zhao, Jian Zhang, Luigi
  Nardi, Peter Bailis, Kunle Olukotun, Chris R{\'e}, and Matei Zaharia.
\newblock Dawnbench: An end-to-end deep learning benchmark and competition.
\newblock \emph{NIPS MLSys}, 2017.

\bibitem[Dauphin et~al.(2017)Dauphin, Fan, Auli, and Grangier]{dauphin2016glu}
Yann~N. Dauphin, Angela Fan, Michael Auli, and David Grangier.
\newblock Language modeling with gated convolutional networks.
\newblock In \emph{International Conference on Machine Learning}, 2017.

\bibitem[Devlin et~al.(2018)Devlin, Chang, Lee, and Toutanova]{devlin2018bert}
Jacob Devlin, Ming-Wei Chang, Kenton Lee, and Kristina Toutanova.
\newblock Bert: Pre-training of deep bidirectional transformers for language
  understanding.
\newblock \emph{arXiv preprint arXiv:1810.04805}, 2018.

\bibitem[Dunn et~al.(2017)Dunn, Sagun, Higgins, Guney, Cirik, and
  Cho]{dunn2017searchqa}
Matthew Dunn, Levent Sagun, Mike Higgins, V~Ugur Guney, Volkan Cirik, and
  Kyunghyun Cho.
\newblock Searchqa: A new q\&a dataset augmented with context from a search
  engine.
\newblock \emph{arXiv preprint arXiv:1704.05179}, 2017.

\bibitem[Gehring et~al.(2017)Gehring, Auli, Grangier, Yarats, and
  Dauphin]{gehring2017convseq}
Jonas Gehring, Michael Auli, David Grangier, Denis Yarats, and Yann~N. Dauphin.
\newblock Convolutional sequence to sequence learning.
\newblock In \emph{International Conference on Machine Learning}, 2017.

\bibitem[Hochreiter and Schmidhuber(1997)]{Hochreiter1997LongSM}
Sepp Hochreiter and J{\"u}rgen Schmidhuber.
\newblock Long short-term memory.
\newblock \emph{Neural computation}, 9 8:\penalty0 1735--80, 1997.

\bibitem[Huang et~al.(2017)Huang, Liu, Van Der~Maaten, and
  Weinberger]{huang2017densely}
Gao Huang, Zhuang Liu, Laurens Van Der~Maaten, and Kilian~Q Weinberger.
\newblock Densely connected convolutional networks.
\newblock In \emph{Proceedings of the IEEE conference on computer vision and
  pattern recognition}, pages 4700--4708, 2017.

\bibitem[Huang et~al.(2018)Huang, Zhu, Shen, and Chen]{huang2018fusionnet}
Hsin-Yuan Huang, Chenguang Zhu, Yelong Shen, and Weizhu Chen.
\newblock Fusionnet: Fusing via fully-aware attention with application to
  machine comprehension.
\newblock In \emph{International Conference on Learning Representations}, 2018.

\bibitem[Joshi et~al.(2017)Joshi, Choi, Weld, and Zettlemoyer]{joshi2017trivia}
Mandar Joshi, Eunsol Choi, Daniel~S. Weld, and Luke Zettlemoyer.
\newblock Triviaqa: A large scale distantly supervised challenge dataset for
  reading comprehension.
\newblock In \emph{Annual Meeting of the Association for Computational
  Linguistics}, 2017.

\bibitem[Kaiser et~al.(2017)Kaiser, Gomez, and Chollet]{kaiser2017depthwise}
Lukasz Kaiser, Aidan~N Gomez, and Francois Chollet.
\newblock Depthwise separable convolutions for neural machine translation.
\newblock \emph{arXiv preprint arXiv:1706.03059}, 2017.

\bibitem[Kingma and Ba(2015)]{kingma2014adam}
Diederik~P Kingma and Jimmy Ba.
\newblock Adam: A method for stochastic optimization.
\newblock In \emph{International Conference on Learning Representations}, 2015.

\bibitem[Kingma et~al.(2015)Kingma, Salimans, and
  Welling]{kingma2015variational}
Diederik~P Kingma, Tim Salimans, and Max Welling.
\newblock Variational dropout and the local reparameterization trick.
\newblock In \emph{Advances in Neural Information Processing Systems}, 2015.

\bibitem[Ko{\v{c}}isk{\`y} et~al.(2018)Ko{\v{c}}isk{\`y}, Schwarz, Blunsom,
  Dyer, Hermann, Melis, and Grefenstette]{kovcisky2018narrativeqa}
Tom{\'a}{\v{s}} Ko{\v{c}}isk{\`y}, Jonathan Schwarz, Phil Blunsom, Chris Dyer,
  Karl~Moritz Hermann, G{\'a}abor Melis, and Edward Grefenstette.
\newblock The narrativeqa reading comprehension challenge.
\newblock \emph{Transactions of the Association of Computational Linguistics},
  6:\penalty0 317--328, 2018.

\bibitem[Kwiatkowski et~al.(2019)Kwiatkowski, Palomaki, Redfield, Collins,
  Parikh, Alberti, Epstein, Polosukhin, Kelcey, Devlin, Lee, Toutanova, Jones,
  Chang, Dai, Uszkoreit, Le, and Petrov]{kwiatkowski2019natural}
Tom Kwiatkowski, Jennimaria Palomaki, Olivia Redfield, Michael Collins, Ankur
  Parikh, Chris Alberti, Danielle Epstein, Illia Polosukhin, Matthew Kelcey,
  Jacob Devlin, Kenton Lee, Kristina~N. Toutanova, Llion Jones, Ming-Wei Chang,
  Andrew Dai, Jakob Uszkoreit, Quoc Le, and Slav Petrov.
\newblock Natural questions: a benchmark for question answering research.
\newblock \emph{Transactions of the Association of Computational Linguistics},
  2019.

\bibitem[Lei et~al.(2017)Lei, Zhang, and Artzi]{lei2017sru}
Tao Lei, Yu~Zhang, and Yoav Artzi.
\newblock Training rnns as fast as cnns.
\newblock \emph{arXiv preprint arXiv:1709.02755}, 2017.

\bibitem[Lei et~al.(2018)Lei, Zhang, Wang, Dai, and Artzi]{lei2018sru}
Tao Lei, Yu~Zhang, Sida~I. Wang, Hui Dai, and Yoav Artzi.
\newblock Simple recurrent units for highly parallelizable recurrence.
\newblock In \emph{Empirical Methods in Natural Language Processing (EMNLP)},
  2018.

\bibitem[McCann et~al.(2017)McCann, Bradbury, Xiong, and
  Socher]{McCann2017CoVe}
Bryan McCann, James Bradbury, Caiming Xiong, and Richard Socher.
\newblock Learned in translation: Contextualized word vectors.
\newblock In \emph{Advances in Neural Information Processing Systems}, 2017.

\bibitem[Miller et~al.(2017)Miller, Feng, Fisch, Lu, Batra, Bordes, Parikh, and
  Weston]{miller2017parlai}
Alexander~H Miller, Will Feng, Adam Fisch, Jiasen Lu, Dhruv Batra, Antoine
  Bordes, Devi Parikh, and Jason Weston.
\newblock Parlai: A dialog research software platform.
\newblock \emph{arXiv preprint arXiv:1705.06476}, 2017.

\bibitem[Paszke et~al.(2017)Paszke, Gross, Chintala, Chanan, Yang, DeVito, Lin,
  Desmaison, Antiga, and Lerer]{paszke2017automatic}
Adam Paszke, Sam Gross, Soumith Chintala, Gregory Chanan, Edward Yang, Zachary
  DeVito, Zeming Lin, Alban Desmaison, Luca Antiga, and Adam Lerer.
\newblock Automatic differentiation in pytorch.
\newblock In \emph{NIPS-W}, 2017.

\bibitem[Pennington et~al.(2014)Pennington, Socher, and
  Manning]{pennington2014glove}
Jeffrey Pennington, Richard Socher, and Christopher~D Manning.
\newblock Glove: Global vectors for word representation.
\newblock In \emph{Empirical Methods in Natural Language Processing}, 2014.

\bibitem[Rajpurkar et~al.(2016)Rajpurkar, Zhang, Lopyrev, and
  Liang]{rajpurkar2016squad}
Pranav Rajpurkar, Jian Zhang, Konstantin Lopyrev, and Percy Liang.
\newblock Squad: 100,000+ questions for machine comprehension of text.
\newblock In \emph{Empirical Methods in Natural Language Processing}, 2016.

\bibitem[Rajpurkar et~al.(2018)Rajpurkar, Jia, and Liang]{rajpurkar2018know}
Pranav Rajpurkar, Robin Jia, and Percy Liang.
\newblock Know what you don’t know: Unanswerable questions for squad.
\newblock In \emph{Proceedings of the 56th Annual Meeting of the Association
  for Computational Linguistics (Volume 2: Short Papers)}, volume~2, pages
  784--789, 2018.

\bibitem[Reddy et~al.(2018)Reddy, Chen, and Manning]{reddy2018coqa}
Siva Reddy, Danqi Chen, and Christopher~D Manning.
\newblock Coqa: A conversational question answering challenge.
\newblock \emph{arXiv preprint arXiv:1808.07042}, 2018.

\bibitem[Seo et~al.(2017)Seo, Kembhavi, Farhadi, and
  Hajishirzi]{seo2016bidirectional}
Minjoon Seo, Aniruddha Kembhavi, Ali Farhadi, and Hannaneh Hajishirzi.
\newblock Bidirectional attention flow for machine comprehension.
\newblock \emph{International Conference on Learning Representations}, 2017.

\bibitem[Seo et~al.(2018)Seo, Min, Farhadi, and Hajishirzi]{seo2017neural}
Minjoon Seo, Sewon Min, Ali Farhadi, and Hannaneh Hajishirzi.
\newblock Neural speed reading via skim-rnn.
\newblock In \emph{International Conference on Learning Representations}, 2018.

\bibitem[Shen et~al.(2018{\natexlab{a}})Shen, Zhou, Long, Jiang, and
  Zhang]{shen2018bi}
Tao Shen, Tianyi Zhou, Guodong Long, Jing Jiang, and Chengqi Zhang.
\newblock Bi-directional block self-attention for fast and memory-efficient
  sequence modeling.
\newblock \emph{arXiv preprint arXiv:1804.00857}, 2018{\natexlab{a}}.

\bibitem[Shen et~al.(2018{\natexlab{b}})Shen, Zhou, Long, Jiang, and
  Zhang]{shen2018fast}
Tao Shen, Tianyi Zhou, Guodong Long, Jing Jiang, and Chengqi Zhang.
\newblock Fast directional self-attention mechanism.
\newblock \emph{arXiv preprint arXiv:1805.00912}, 2018{\natexlab{b}}.

\bibitem[Srivastava et~al.(2014)Srivastava, Hinton, Krizhevsky, Sutskever, and
  Salakhutdinov]{Srivastava2014dropout}
Nitish Srivastava, Geoffrey Hinton, Alex Krizhevsky, Ilya Sutskever, and Ruslan
  Salakhutdinov.
\newblock Dropout: A simple way to prevent neural networks from overfitting.
\newblock \emph{Journal of Machine Learning Research}, 2014.

\bibitem[Vaswani et~al.(2017)Vaswani, Shazeer, Parmar, Uszkoreit, Jones, Gomez,
  Kaiser, and Polosukhin]{vaswani2017attention}
Ashish Vaswani, Noam Shazeer, Niki Parmar, Jakob Uszkoreit, Llion Jones,
  Aidan~N Gomez, {\L}ukasz Kaiser, and Illia Polosukhin.
\newblock Attention is all you need.
\newblock In \emph{Advances in Neural Information Processing Systems}, 2017.

\bibitem[Vinyals et~al.(2015)Vinyals, Fortunato, and
  Jaitly]{vinyals2015pointer}
Oriol Vinyals, Meire Fortunato, and Navdeep Jaitly.
\newblock Pointer networks.
\newblock In \emph{Advances in Neural Information Processing Systems}, 2015.

\bibitem[Wu et~al.(2017)Wu, Lao, Blitzer, Yang, and Weinberger]{wu2017fast}
Felix Wu, Ni~Lao, John Blitzer, Guandao Yang, and Kilian Weinberger.
\newblock Fast reading comprehension with convnets.
\newblock \emph{arXiv preprint arXiv:1711.04352}, 2017.

\bibitem[Wu et~al.(2019)Wu, Fan, Baevski, Dauphin, and Auli]{wu2018pay}
Felix Wu, Angela Fan, Alexei Baevski, Yann Dauphin, and Michael Auli.
\newblock Pay less attention with lightweight and dynamic convolutions.
\newblock In \emph{International Conference on Learning Representations}, 2019.
\newblock URL \url{https://openreview.net/forum?id=SkVhlh09tX}.

\bibitem[Yang et~al.(2015)Yang, Yih, and Meek]{yang2015wikiqa}
Yi~Yang, Wen-tau Yih, and Christopher Meek.
\newblock Wikiqa: A challenge dataset for open-domain question answering.
\newblock In \emph{Proceedings of the 2015 Conference on Empirical Methods in
  Natural Language Processing}, pages 2013--2018, 2015.

\bibitem[Yu et~al.(2017)Yu, Lee, and Le]{yu2017learning}
Adams~Wei Yu, Hongrae Lee, and Quoc~V Le.
\newblock Learning to skim text.
\newblock In \emph{Annual Meeting of the Association for Computational
  Linguistics}, 2017.

\bibitem[Yu et~al.(2018)Yu, Dohan, Le, Luong, Zhao, and Chen]{wei2018qanet}
Adams~Wei Yu, David Dohan, Quoc Le, Thang Luong, Rui Zhao, and Kai Chen.
\newblock Qanet: Combining local convolution with global self-attention for
  reading comprehension.
\newblock In \emph{International Conference on Learning Representations}, 2018.

\bibitem[Zhang et~al.(2015)Zhang, Zhao, and LeCun]{zhang2015character}
Xiang Zhang, Junbo Zhao, and Yann LeCun.
\newblock Character-level convolutional networks for text classification.
\newblock In \emph{Advances in neural information processing systems}, 2015.

\end{thebibliography}
